\documentclass[conference]{IEEEtran}
\IEEEoverridecommandlockouts
\usepackage{cite}
\usepackage{amsmath,amssymb,amsfonts}
\usepackage{algorithmic}
\usepackage{graphicx}
\usepackage{subcaption}
\usepackage{tabularray}
\usepackage{textcomp}
\usepackage{xcolor}
\usepackage{booktabs}
\usepackage{multirow}
\usepackage{hhline}
\usepackage{makecell}
\usepackage{threeparttable}
\usepackage{authblk}
\usepackage{tcolorbox}

\def\BibTeX{{\rm B\kern-.05em{\sc i\kern-.025em b}\kern-.08em
    T\kern-.1667em\lower.7ex\hbox{E}\kern-.125emX}}
    
\newcommand\ourframework{\textsc{QuLog}}    

\begin{document}

\title{Quantum Machine Learning in Log-based Anomaly Detection: Challenges and Opportunities}

\author{Jiaxing Qi}
\author{Chang Zeng}
\author{Zhongzhi Luan\thanks{Corresponding author: luan.zhongzhi@buaa.edu.cn}}
\author{Shaohan Huang}
\author{Shu Yang} 
\author{\\ Yao Lu}
\author{Bin Han}
\author{Hailong Yang}
\author{Depei Qian}

\affil{Sino-German Joint Software Institute, Beihang University, Beijing, China}


\maketitle

\begin{abstract}
Log-based anomaly detection (LogAD) is the main component of Artificial Intelligence for IT Operations (AIOps), which can detect anomalous that occur during the system on-the-fly. Existing methods commonly extract log sequence features using classical machine learning techniques to identify whether a new sequence is an anomaly or not. However, these classical approaches often require trade-offs between efficiency and accuracy. The advent of quantum machine learning (QML) offers a promising alternative. By transforming parts of classical machine learning computations into parameterized quantum circuits (PQCs), QML can significantly reduce the number of trainable parameters while maintaining accuracy comparable to classical counterparts. 
%
In this work, we introduce a unified framework, \ourframework{}, for evaluating QML models in the context of LogAD. This framework incorporates diverse log data, integrated QML models, and comprehensive evaluation metrics. State-of-the-art methods such as DeepLog, LogAnomaly, and LogRobust, along with their quantum-transformed counterparts, are included in our framework.
%
Beyond standard metrics like F1 score, precision, and recall, our evaluation extends to factors critical to QML performance, such as specificity, the number of circuits, circuit design, and quantum state encoding. Using \ourframework{}, we conduct extensive experiments to assess the performance of these models and their quantum counterparts, uncovering valuable insights and paving the way for future research in QML model selection and design for LogAD.
\end{abstract}


\section{Introduction}

As the complexity and scale of IT systems continue to grow, the operation and maintenance capabilities of Internet Service Providers (ISPs) are facing significant challenges. Gartner \cite{abbas2024aiops} proposed the concept of Artificial Intelligence for IT Operations (AIOps), which involves collecting and analyzing system data using big data and artificial intelligence technologies to inform operation and maintenance decisions. Log-based anomaly detection (LogAD) is a core capability in AIOps. Particularly in software-intensive systems, it can help the operator to automatically detect faults, which is crucial for the high availability and reliability of the system \cite{landauer2024critical}. 
\par
Over the past years, LogAD has attracted significant attention from both researchers and industry community. Given the large-scale, semi-structured, high-dimensional, and noisy nature of log data, many machine learning-based models have been proposed (see Section \ref{sec: logad} for more details) \cite{le2022log}. According to training strategies, existing models can be grouped into three categories: supervised, semi-supervised, and unsupervised. Supervised-based methods are usually considered as a classification task, such as LogRobust \cite{zhang2019robust}. Although it can achieve good accuracy, it requires sufficient labeled log data, which typically demands considerable human effort. Semi-supervised methods require only normal log data, such as LogAnomaly \cite{meng2019loganomaly}, and learn the system's normal behavior pattern to identify anomalies in new log data. These methods have become mainstream due to their accuracy and low human effort. Unsupervised methods do not require prior log information and use clustering and association analysis techniques, such as LogCluster \cite{vaarandi2015logcluster}, yet often have poor accuracy because of limited domain knowledge. In a nutshell, achieving better model accuracy requires more human effort and more complex model designs, which in turn necessitate additional log data. ISPs must make a trade-off between efficiency and accuracy.

\begin{figure}
    \centering
    \includegraphics[scale=0.5]{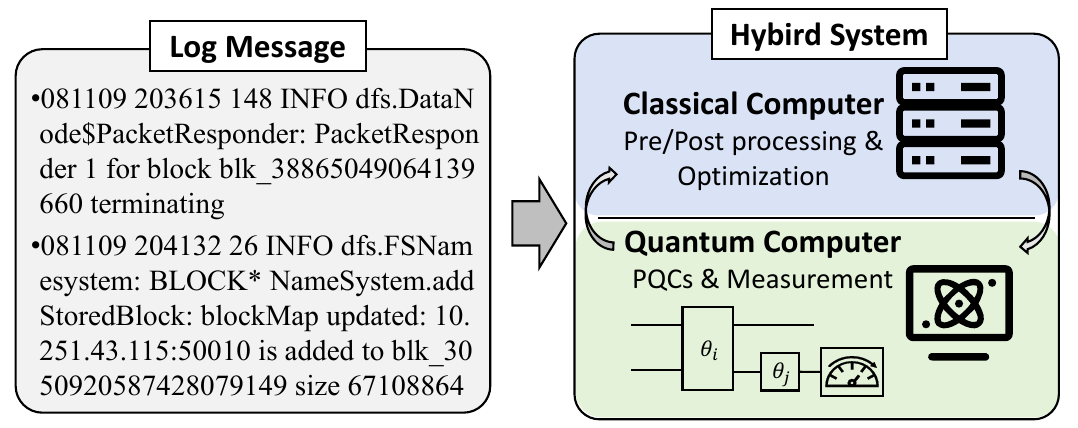}
    \caption{High-level depiction of hybrid quantum machine learning system for log analysis.}
    \label{fig:intro}
\end{figure}

\begin{figure}
    \centering
    \includegraphics[scale=0.6]{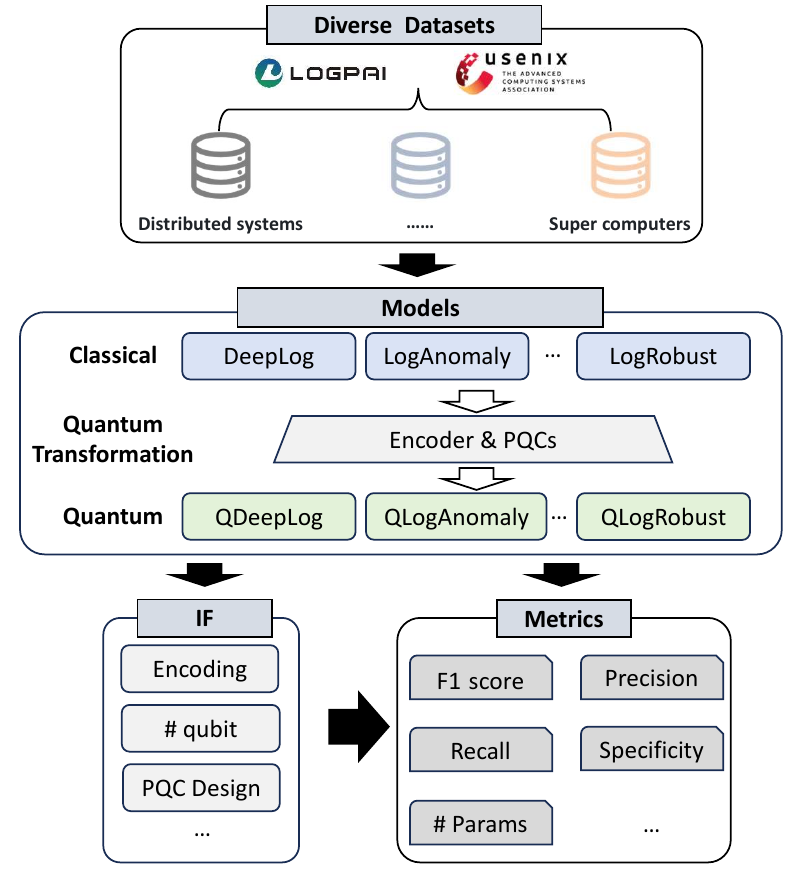}
    \caption{Overview of the \ourframework{} framework. The framework is structured incrementally, progressing from datasets to models and then to metrics.}
    \label{fig:framework}
\end{figure}

\par
Given above landscape, we introduce Quantum Machine Learning (QML) into LogAD. QML is a cutting-edge technology that combines quantum computing and machine learning and has shown significant potential and promising outcomes in processing large-scale, high-dimensional data. The advantage is leveraging the superposition and entanglement properties of quantum bits (qubits) to implement parameterized quantum circuits (PQCs), which theoretically enables the efficient handling and pattern recognition of complex data structures. This capability substantially surpasses the processing power of classical computers, particularly in solving specific optimization problems and conducting high-dimensional data analysis (see Section \ref{sec: qml} for more details).
\par
To explore the potential of QML in LogAD, we first propose a unified framework, named \ourframework{}, which incorporates diverse log data, integrated QML models, and comprehensive evaluation metrics. Figure \ref{fig:framework} provides an overview of the framework, and a summary of each component is as follows:

\begin{itemize}
    \item \textbf{Diverse datasets:} We have collected rich log datasets from LogPai's LogHub repository\footnote{https://github.com/logpai/loghub} and the Usenix website\footnote{https://www.usenix.org/cfdr-data\#hpc4}, covering various areas such as distributed systems and supercomputers. These datasets not only provide researchers with valuable resources to develop and evaluate advanced log analysis techniques, but also support application research of QNNs on large datasets.
    \item \textbf{QML models:} We design and implement a set of hybrid machine learning models based on quantum-classical computing by innovatively ``quantumizing" popular LogAD models, including DeepLog, LogAnomaly, and LogRobust. These hybrid ML models not only retain the core logic of their classical counterparts but also significantly reduce the model's parameter scale through qubit parallelism and quantum gate operations.
    \item \textbf{Comprehensive evaluation metric:}  In addition to commonly used metrics such as F1 score, precision, and recall, we also incorporated Specificity to provide a more comprehensive evaluation of the model's performance. Furthermore, we evaluated the model's parameter size, which is essential for deploying the model on resource-limited devices.
\end{itemize}

Through comprehensive experimental evaluations, we compared the performance of classical machine learning models with hybrid QML models and identified the following key findings:
\begin{itemize}
    \item \textbf{Dataset characteristics.} The variation in performance across different datasets highlights the importance of dataset characteristics in QML applications and suggests that QML model design may not generalize to different log data.
    \item \textbf{Quantum encoding and circuit design.} Quantum circuit design, including factors such as the number of qubits, the choice of unitary gates, and circuit complexity, significantly influences performance. Simpler circuits are often more effective than overly complex ones. Additionally, transforming classical models into quantum variants does not yield optimal results; instead, quantum models should be designed from scratch.
    \item \textbf{Robustness to training set size.} Quantum models demonstrated robustness in maintaining high recall across different training set sizes. However, they exhibited oscillations in precision and specificity, indicating room for further optimization.
    \item \textbf{Training efficiency.} The stable and efficient convergence observed in most QML models highlights the effectiveness of quantum optimization algorithms, such as parameter-shift methods, in minimizing loss functions. Training stability can be further improved through circuit simplification or refined initialization strategies.
\end{itemize}

\par
The major contributions of this work are as follows:

\begin{itemize}
    \item We first introduced quantum machine learning to log-based anomaly detection task and designed a unified framework for evaluating the performance of QML in this task.
    \item We quantumized three commonly used classical models and performed extensive evaluations within our designed framework. Our findings conclude that quantum machine learning is promising for log-based anomaly detection.
    \item Based on the evaluation results, we summarised the advantages and disadvantages of QML applied to log-based anomaly detection and proposed future research directions.
\end{itemize}

\section{Preliminary}

\subsection{Log-based Anomaly Detection} \label{sec: logad}
Log-based Anomaly Detection (LogAD) task has been extensively studied in existing literature, we will not introduce it in detail in our work. Instead, we provide a brief summary of the general process for this task. It consists of the following four steps: 
\par
\textbf{1) Log preprocessing.} Log data is generally considered semi-structured, consisting of a program-defined event description (constant part) and a dynamically changing part that varies based on the system's operating state (variable part). To automatically parse log data to extract log events and variables, many log parsing methods have been proposed \cite{zhu2019tools}. The most commonly used method is Drain \cite{he2017drain} due to its efficiency and accuracy. Then, a fixed-size sliding time window is employed to group the logs into log sequences. Depending on the type of log data within the sequence, there are three categories: raw log sequence, log content sequence, and log event sequence. Typically, we use the log event sequence for subsequent anomaly detection according to previous work settings \cite{li2024graph}. Finally, this step may also address missing or duplicate data in the log sequence. For supervised methods \cite{guo2024logformer}, to address the imbalance between positive and negative log samples, the abnormal log sequence is often over-sampled, for example, by injecting anomalies into the log sequence.
\par
\textbf{2) Log vectorization.} In this step, we need to transform the log sequence into a dense, real-valued vector representation as the input for the machine learning model. Common approaches, including LogAnomaly and LogRobust, consider each log record as natural language and extract the semantic information of each log using Word2Vec \cite{mikolov2013distributed} and BERT embeddings \cite{devlin2018bert} \cite{guo2021logbert}. Notably, DeepLog uses one-hot encoding for each event, while LogAnomaly incorporates an additional event quantitative vector.
\par
\textbf{3) LogAD modeling.} After the log sequence is vectorized, we train the machine learning model to detect anomalies within each sequence. Common models used include SVM \cite{yu2024deep}, RNN \cite{zaremba2014recurrent}, CNN \cite{lu2018detecting}, K-means \cite{huang2024application}, and Transformer \cite{vaswani2017attention}, etc. Based on the training algorithms, these models can be categorized into supervised, semi-supervised, and unsupervised. Supervised-based models, such as SVM and CNN, require a labeled dataset for training. These models learn to distinguish between normal and anomalous sequences by analyzing examples from both classes. While supervised models often yield high accuracy when provided with sufficient labeled data, their performance can degrade in scenarios where labeled anomalies are sparse or difficult to obtain in practice. Semi-supervised-based models, like variants of RNNs \cite{wang2021multi} and Autoencoders \cite{catillo2022autolog}, are trained primarily on normal sequences, assuming that anomalous events are rare or not well-represented in the training set. The model learns to reconstruct or predict normal behavior, and sequences deviating significantly from the learned patterns are identified as anomalies. These models strike a balance between human effort and robustness.
\par
\textbf{4) Evaluation.} Selecting and evaluating models requires the use of appropriate performance metrics. Common metrics include precision, recall, and F1 score. In scenarios where anomalies are extremely rare and recall is high, specificity often provides more meaningful insights than accuracy. Additionally, model size plays a critical role, particularly for deployment on resource-constrained devices.

\subsection{Quantum Machine Learning} \label{sec: qml}

\begin{figure}[!tb]
    \centering
    \includegraphics[scale=0.5]{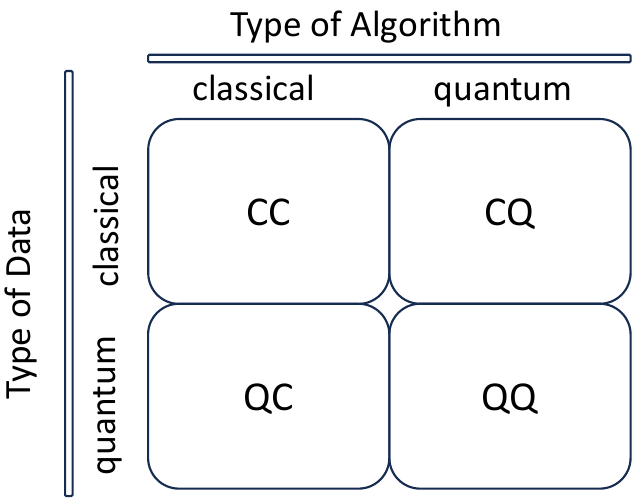}
    \caption{Four types of combine the disciplines of quantum computing and machine learning.}
    \label{fig:qml_desc}
\end{figure}

Machine Learning (ML) aims to enable systems to learn from data, recognize patterns, and make decisions without explicit programming. ML algorithms, however, face computational limitations, especially with large datasets and complex computations, as they scale in classical, often Euclidean space \cite{gil2024understanding, jerbi2024shadows}. Quantum Machine Learning (QML) emerges as an extension of ML, leveraging principles of quantum mechanics to potentially solve these challenges more efficiently. QML could provide advantages such as faster data processing, more efficient memory usage, and the ability to explore larger solution spaces due to quantum superposition and entanglement \cite{cerezo2022challenges}.
\par
The difference between ML and QML can be summarized as follows: 
\begin{itemize}
    \item Classical ML operates in Euclidean space, represented by linear transformations on vector spaces.
    \item QML, on the other hand, operates in a complex Hilbert space, exploiting superposition and entanglement to encode and process information in non-Euclidean space.
\end{itemize}
\par
QML combines classical ML techniques with quantum computing principles, a broad term encompassing various approaches to integrate quantum computing and ML disciplines (see Figure \ref{fig:qml_desc}) \cite{jerbi2023quantum}. There are four types of tasks combining different data and algorithm types. For example, one can train ML models with classical algorithms using classical data, which has been extensively applied in image and text processing (top-left), or use quantum algorithms to train ML models for processing either classical or quantum data (right side). Even classical tasks may be regarded as QML if they are quantum-inspired. We note that the focus of this paper will be on quantum neural networks, which function as models similar to classical neural networks but utilize quantum bits (qubits) and quantum gates for computation. However, the field of QML is quite broad and extends beyond these topics.
\par

Here, we begin by introducing several key components essential for implementing quantum neural network (QNN), including data encoding, unitary operation, parameterized quantum circuits (PQCs), and measurement.


\subsubsection{Data Encoding} Classical data is encoded in bits. In classical computers, each bit has two states, 0 or 1. For example, images, text, and log data are all encoded in bits. However, QNNs can only process data encoded as quantum bits (qubits). A quantum bit is represented as $|0\rangle$ or $|1\rangle$, or any normalized complex linear \textit{superposition} of these two. Therefore, it is necessary to encode classical data into quantum data, that is, to encode it using qubits. Existing encoding methods include amplitude, angle encoding, and basis encoding \cite{rath2024quantum}\cite{biamonte2017quantum}\cite{tian2023recent}. Once encoded into quantum data, it is already formalized as a set of quantum states $\{|\varphi_i\rangle\}$ or a set of unitaries $\{U_i\} $ that can be prepared on a quantum device via the relation $|\varphi_j = U_j |0\rangle$. For example, given a classical data $x_i$, the encoding process can be formalized in Eq.\ref{eq:data_encode}. In this way, QML maps the data into the Hilbert space of exponentially large dimensions to solve the learning task.

\begin{equation}\label{eq:data_encode}
    \begin{array}{cc}
         &  x_i \rightarrow |\varphi(x_i)\rangle\\
    \end{array}
\end{equation}
\par
where the $|\varphi(x_i)\rangle$ is in a Hilbert space $\mathcal{H}$. 

\begin{figure}
    \centering
    \includegraphics[scale=0.4]{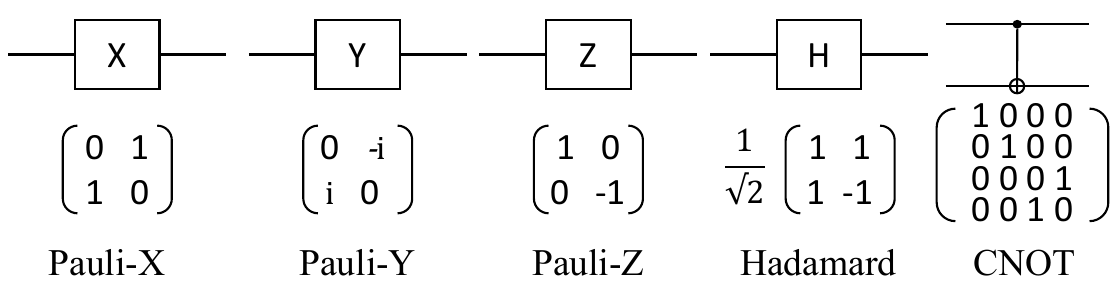}
    \caption{Commonly used quantum gates and their corresponding unitary matrices.}
    \label{fig:gate}
\end{figure}
\subsubsection{Unitary Operation (Quantum Gate)} In QML, transformations are represented by unitary operations, implemented via quantum gates. Unlike classical operations, these are reversible and ensure no information loss. Common quantum gates include \textit{Pauli} gates, \textit{Hadamard} gates, and \textit{controlled-NOT} (CNOT) gates et.al \cite{rath2024quantum} (see Figure \ref{fig:gate}). Each gate performs a specific type of transformation, affecting the quantum state based on probabilistic principles. 

\begin{figure}
    \centering
    \includegraphics[scale=0.35]{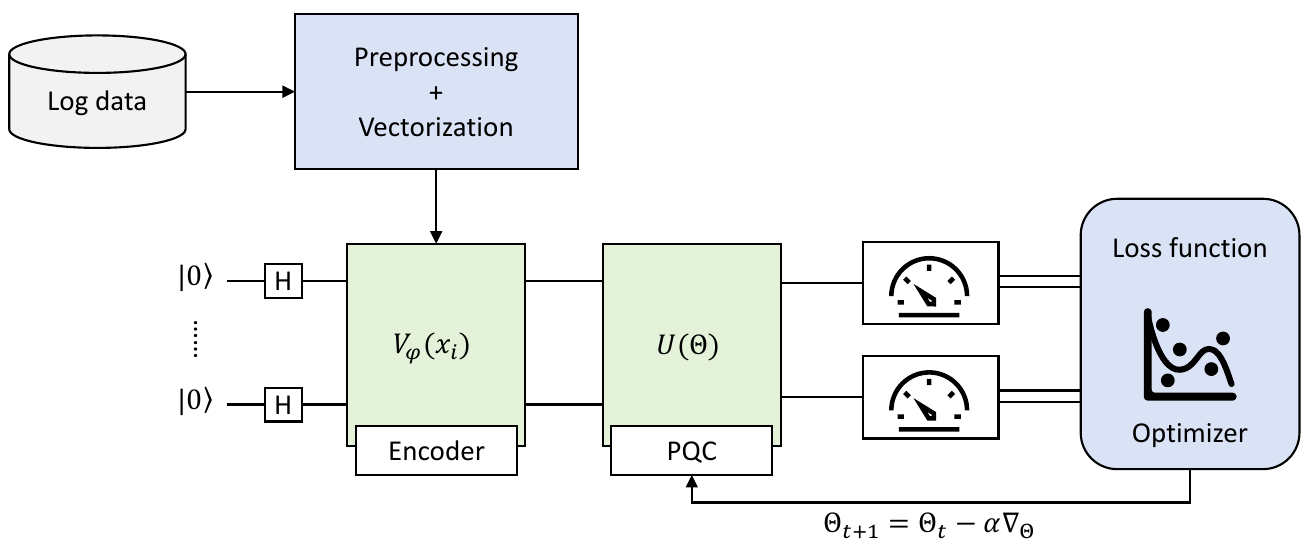}
    \caption{The training process of the classical-quantum hybrid machine learning model for log-based anomaly detection. The purple module represents the classical computer operation and the green part represents the quantum computer operation.}
    \label{fig:pqc}
\end{figure}

\subsubsection{Quantum Circuits} A quantum circuit is a sequence of quantum gates applied to qubits that enables complex transformations on the encoded quantum states. In QNNs, quantum circuits are designed to include specific configurations of quantum gates, often called layers, that can manipulate qubits to represent more intricate patterns in data. The arrangement and choice of gates in a quantum circuit are critical, as they determine the expressiveness and efficiency of the QNN model. Quantum circuits can be fixed, or they may contain tunable parameters, allowing them to act as trainable components, much like layers in a classical neural network.
\par
Quantum circuits used in QNNs are typically structured as Parameterized Quantum Circuits (PQC), which include gates with adjustable parameters. These parameters are optimized during the learning process, enabling the QNN to “learn” from data and adapt the model to minimize a defined loss function. PQCs play a central role in the QNN architecture and contribute to its ability to generalize across various types of data. 

\subsubsection{Measurement} Once the data has been processed by the quantum circuit, a measurement step is performed to extract classical information from the quantum states. Measurement collapses the superposition state of each qubit into one of its basis states (e.g., $|0\rangle$ or $|1\rangle$) based on their probability amplitudes. This final measurement result yields the output that can be used for further processing or interpreted directly as a solution to the learning task. 
\par
Measurement in QML is inherently probabilistic, so multiple measurements are often needed to obtain statistically meaningful results. The outcome probabilities reflect the information encoded in the final state of the quantum circuit, giving a classical interpretation of the quantum computation performed. These measured results can then be used to update the PQC parameters during the training phase, iteratively refining the QNN’s performance.

\subsection{Training Process of QNNs} The training of QNNs generally follows a similar process to classical neural networks, where parameters are optimized to minimize a loss function. However, due to the quantum nature of the circuits, QNN training often employs a hybrid approach. This process is depicted in Figure \ref{fig:pqc}.

\begin{figure*}
    \centering
    \includegraphics[scale=0.6]{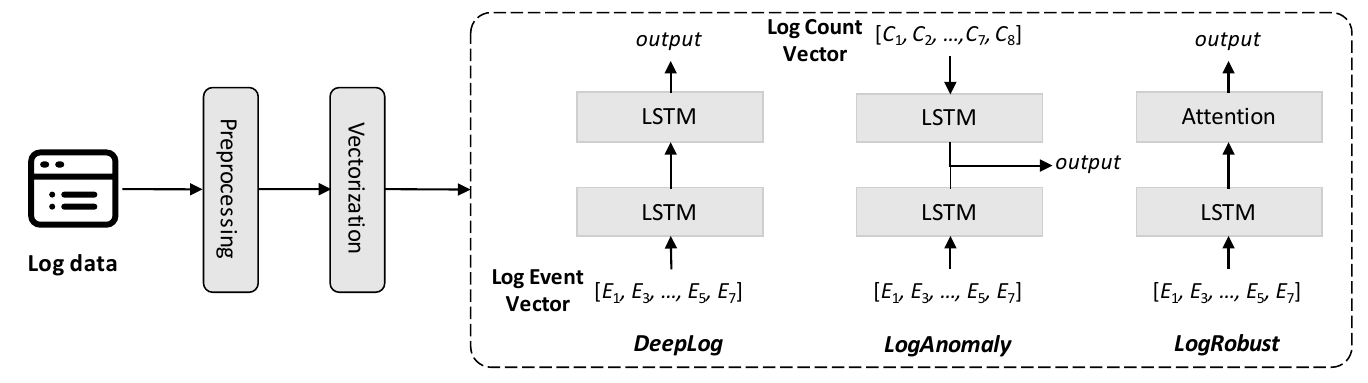}
    \caption{The workflow of classical baseline methods. We overlook the preprocessing and vectorization step, and the three baselines primarily utilize two type models: LSTM and Attention.}
    \label{fig:baseline}
\end{figure*}

\section{Baseline Models \& Quantum Transformation}
We selected commonly used classical baselines, including DeepLog, LogAnomaly, and LogRobust, and performed quantum transformation on their key computational steps. Specifically, we carefully designed PQCs to replace the classical computational processes, while retaining some classical computation steps. Consequently, the transformed model is a classical-quantum hybrid model.

\subsection{Classical Baseline Models}  \label{sec:baseline}
First, we describe the selected baseline methods and analyze their computational steps. The workflow of three methods is depicted in Figure \ref{fig:baseline}.
\begin{itemize}
    \item \textbf{DeepLog} \cite{du2017deeplog} treats logs as a time-series sequence, using the current sequence of events to predict the next possible log event. It leverages an LSTM model to automatically extract patterns of normal operations. When the observed log patterns deviate from the trained model, anomalies can be detected.
    \item \textbf{LogAnomaly} \cite{meng2019loganomaly} introduced a log count vector to train an LSTM model, capturing the dependence among events in a log sequence. Additionally, it designed \textit{template2Vec}, a semantic vector representation of log events generated by building a synonym-antonym dictionary, allowing it to match unseen events based on existing events. Finally, similar to DeepLog, LogAnomaly trains an LSTM model using a sequence of normal logs as input to predict the next likely event. An event sequence is considered normal only when both LSTM models' predictions match the actual output; otherwise, it is identified as anomalous.
    \item \textbf{LogRobust} \cite{zhang2019robust} addresses the instability of logs, meaning that log events evolve over time, and noise may exist within log data. To overcome this, it represents logs as semantic vectors by combining a pre-trained word2vec model with TF-IDF weights. It then leverages an attention-based Bi-LSTM model to detect anomalies, capturing contextual information in log sequences and automatically assessing the importance of different log events. As a result, LogRobust effectively identifies and solves unstable log events.
\end{itemize}
\par
In summary, the models used in the three baseline methods can be categorized into two types: LSTM model and self-attention model. Consequently, the problem of quantumizing these methods reduces to determining \textbf{``How to quantumized the LSTM and Attention models?"}.
\subsection{Quantum Transformation}
To quantumized the LSTM and Attention models for use in the quantum framework, it is necessary to understand their structural and operational principles and adapt them for quantum computation \cite{chen2022quantum, vaswani2017attention}. This process involves mapping classical components of these models onto quantum counterparts that can leverage quantum properties such as superposition and entanglement to potentially enhance performance and scalability.
\par
A classic LSTM cell primarily comprises three gates: the forget gate, the input gate, and the output gate. After the input is processed by these three gates, the final outputs, $C_t$ and $h_t$, are obtained. The main calculations are formalized as follows:
\begin{equation}  \label{eq:lstm_cell}
    \begin{array}{cc}
         \textbf{forget gate: } & f_t = \sigma (W_f \cdot [h_{t-1}, x_t] + b_f) \\ 
         \textbf{input gate: }  & \\
         &   i_t = \sigma(W_i \cdot [h_{t-1}, x_t] + b_i) \\
         &  \tilde{C_t} = tanh(W_C \cdot [h_{t-1}, x_t] + b_C) \\
         &   C_t = f_t * C_{t-1} + i_t * \tilde{C}_{t} \\
         \textbf{output gate: } & o_t = \sigma(W_o\cdot[h_{t-1}, x_t] + b_o) \\
         &  h_t = o_t * tanh(C_t)
    \end{array}
\end{equation}
\par
where $h_t$ and $x_t$ are the hidden state and input data at timestep $t$, $W$ is the trainable parameters of each gate, and $\sigma$ is the \textit{sigmoid} activation function.

\par
The self-attention model, which primarily comprises three different linear transformers: $W_Q$, $W_K$, $W_V$ applies to each input feature vector to transform them to new internal representation called Query ($Q$), Key ($K$), and Value ($V$). These states are then passed to the function that calculates the attention weights. Given a matrix with $k$ inputs $X \in \mathbb{R}^{k\times d}$, the main calculations are formalized as follows:
\begin{equation}
    \begin{array}{cc}
         &  Attention(Q, K, V) = softmax(\frac{QK^T}{\sqrt{d_k}})V \\
         &  Q = XW_Q; K=XW_K; V=XW_V
    \end{array}
\end{equation}

Our goal is to realize the computations of LSTM and self-attention mechanisms through PQCs. Similarly to \cite{chen2022quantum}, we primarily designed and developed two primary modules: the Encoder and PQC, as shown in Figure \ref{fig:pqc}. 
\par
First, the log data is preprocessed, vectorized, and formalized as follows:
\begin{equation}
    \begin{array}{cc}
         &  \{E_t, E_{t+1}, \cdots, E_{t+k-1} \}= \mathcal{P}(\{M_{t}, M_{t+1}, \cdots, M_{t+k-1}\}) \\
         &  \{x_t, x_{t+1}, \cdots, x_{t+k-1} \} = \mathcal{V}(\{E_t, E_{t+1}, \cdots, E_{t+k-1}\})
    \end{array}
\end{equation}
\par
where $M_{i}$ is the $t$th log message, $E_t$ is the event corresponding to $M_{i}$, and $x_t \in \mathbb{R}^d$ is the feature vector as the model input. 
\par
To train the model end-to-end, each feature must be assigned a qubit. Since current quantum computers are limited in the number of qubits $n$ they can provide ($n << d$), we use a linear transformation matrix to downsample the original features (see Eq.\ref{eq:trans_qubit}), enabling the quantum computer to process.
\begin{equation}\label{eq:trans_qubit}
    \begin{array}{cc}
         &  \tilde{x}_t =  x_t \cdot \mathbf{W} + \mathbf{b}\\
    \end{array}
\end{equation}
\par
where $\mathbf{W} \in \mathbb{R}^{d\times n}$ and $\mathbf{b} \in \mathbb{R}^{n}$ are trainable parameters.

\par
\textbf{Encoder.}  Encoding methods aim to represent classical data $\tilde{x}_t$ as quantum states $|\varphi(\tilde{x}_t)\rangle$, enabling operations in a high-dimensional Hilbert space. These states can be prepared on quantum hardware using unitary operators $V_\varphi$:
\begin{equation}
    \begin{array}{cc}
         & |\varphi(\tilde{x}_t)\rangle = V_\varphi |0 \rangle \\
    \end{array}
\end{equation}
\par
Here, $V_\varphi$ represents the quantum circuit that transforms the initial state $|0\rangle$into the encoded state $|\varphi(\tilde{x}_t)\rangle$, corresponding to the input feature $\tilde{x}_t$.  

\par
Specifically, the $H$ gate is used to prepare the unbiased superposition state of the initial state, the preparation of a single qubit is as follows: 
\begin{equation}
    \begin{array}{cc}
         & H|0\rangle = \frac{1}{\sqrt{2}}| (|0\rangle + |1\rangle) = |+\rangle \\
    \end{array}
\end{equation}

\par
Then, we utilize three variants of angle encoding and one type of amplitude encoding. Each encoding method leverages different aspects of the feature data to ensure effective representation in the quantum domain. The mathematical formulations for these specific encoding strategies are as follows.
\par
\textbf{Angle encoding around the X-Axis ($R_x$}). The state of a single qubit is rotated using the $R_{x}$ gate:
\begin{equation}
    \begin{array}{cc}
         &  R_x(\tilde{x}_t[i]) = exp(-i\frac{\tilde{x}_t[i]}{2}\sigma_x) \\
    \end{array}
\end{equation}
where $\sigma_x$ refer to the \textit{Pauli-X} gate. Resulting in the quantum state: 
\begin{equation}
    \begin{array}{cc}
         &  |\varphi(\tilde{x}_t[i])\rangle = cos(\tilde{x}_t[i]/2)|0\rangle + i\cdot sin(\tilde{x}_t[i]/2)|1\rangle \\
    \end{array}
\end{equation}
\par
\textbf{Angle encoding around the Y-Axis ($R_y$}). The state of a single qubit is rotated using the $R_{y}$ gate:
\begin{equation}
    \begin{array}{cc}
         &  R_y(\tilde{x}_t[i]) = exp(-i\frac{\tilde{x}_t[i]}{2}\sigma_y) \\
    \end{array}
\end{equation}
\par
where $\sigma_y$ refer to the \textit{Pauli-Y} gate. Resulting in the quantum state: 
\begin{equation}
    \begin{array}{cc}
         &  |\varphi(\tilde{x}_t[i])\rangle = cos(\tilde{x}_t[i]/2)|0\rangle + sin(\tilde{x}_t[i]/2)|1\rangle \\
    \end{array}
\end{equation}
\par
\textbf{Angle encoding around the Z-Axis ($R_z$}). The state of a single qubit is rotated using the $R_{z}$ gate:
\begin{equation}
    \begin{array}{cc}
         &  R_z(\tilde{x}_t[i]) = exp(-i\frac{\tilde{x}_t[i]}{2}\sigma_z) \\
    \end{array}
\end{equation}
\par
where $\sigma_y$ refer to the \textit{Pauli-Z} gate. Resulting in the quantum state: 
\begin{equation}
    \begin{array}{cc}
         &  |\varphi(\tilde{x}_t[i])\rangle = e^{-i\tilde{x}_t[i]/2}|0\rangle + e^{i\tilde{x}_t[i]/2} |1\rangle \\
    \end{array}
\end{equation}
\par
\textbf{Amplitude encoding}. The state of a single qubit is mapping into the amplitude of a quantum state. The resulting state is represented as:
\begin{equation}
    \begin{array}{cc}
         &  |\varphi(\tilde{x}_t)\rangle = \frac{1}{||\tilde{x}_t||} \sum_{i=1}^{n}\tilde{x}_t[i]|i\rangle\\
    \end{array}
\end{equation}
\par
where $||\tilde{x}_t||$ is the norm of the vector, ensuring normalization of the quantum state. Amplitude encoding is efficient in terms of qubits, as $n$ features are encoded into $log_2(n)$ qubits.
\par
\textbf{PQC Design.}  

Classical data has been encoded into quantum states. Subsequently, these quantum states undergo a series of unitary operations ($U(\Theta)$), including the rotation gate ($R_x, R_y, R_z$) and CNOT gate. By iteratively combining and configuring these gates, various PQCs can be constructed. Specifically, a rotation gate operates on a single quantum bit. Adjusting the parameters of a rotation gate allows control over the amplitude or phase of a single quantum bit. 
\par

In addition, the CNOT gate is a two-qubit operation that establishes non-local correlations between qubits through entanglement, which is one of the advantages of quantum computing over classical computing. The CNOT gate flips the second qubit (the target qubit) if and only if the first qubit (the control qubit) is $|1\rangle$.

By alternating rotation and CNOT gates across multiple layers (see Eq.\ref{eq:pqc}), this PQC is able to generate complex quantum states, thereby enabling highly nonlinear transformations of data.

\begin{equation}\label{eq:pqc}
    \begin{array}{cc}
         &  U(\Theta) = \{R_x(\theta_x), R_y(\theta_y), R_z(\theta_z), \text{CNOT}\}^{*} |\varphi(\tilde{x}_t)[i]\rangle\\
    \end{array}
\end{equation}
\par
where $\{\cdot\}^*$ refer to Kleene closure, $\theta_x$, $\theta_y$, and $\theta_z$ are trainable parameters that rotation angle in terms of X-Axis, Y-Axis, and Z-Axis.
\par
\textbf{Measurement and Optimizer.} After processing through PQCs, the quantum states are measured and the final quantum states collapse into classical information that can be used for predictions or post-processing. Due to the probabilistic nature of quantum measurements, these results are typically averaged over multiple runs to provide reliable outcomes. In this work, the measurement method is beyond the scope of our study, and we refer to \cite{joshi2021evaluating} for its implementation. 
\par
For the optimizer, we employ the parameter-shift method to compute the gradients of PQC parameters with respect to the objective function. For example, given the expectation value of an observable $\tilde{O}$ with encoded quantum state $|\varphi(\tilde{x}_t)\rangle$.
\begin{equation}
    \begin{array}{cc}
         &  f(|\varphi(\tilde{x}_t)\rangle, \Theta) = \langle\varphi(\tilde{x}_t)| U^{\dag}(\Theta)  \tilde{O}  U(\Theta)  |\varphi(\tilde{x}_t)\rangle\\
    \end{array}
\end{equation}
\par
According to the Parameter-Shift Rule \cite{wierichs2022general}, each parameter in a parameterized gate operates on the circuit in the form $e^{-i \frac{\theta_i}{2}P}$, where $P$ is a \textit{Pauli} operator, and the parameterized gate exhibits periodicity with respect to the parameter $\theta_i$ (usually with a period of $2\pi$). The single-qubit operation associated with the parameter $\theta_i$ can derived as follows:
\begin{equation}
     U(\Theta)=U_{\mathrm{pre}} \cdot e^{-i \frac{\theta_i}{2} P} \cdot U_{\mathrm{post}} \\
\end{equation}
\par
where $U_{\mathrm{pre}}$ and $U_{\mathrm{post}}$ are refer to the circuit before and after $e^{-i \frac{\theta_i}{2}P}$. Thus, the $f$ can be reformulated as follows:
\begin{equation}
     f(\Theta)=\langle\psi| U_{\mathrm{pre}}^{\dagger} \cdot e^{i \frac{\theta_i}{2} P} \cdot U_{\mathrm{post}}^{\dagger} \cdot \hat{O} \cdot U_{\mathrm{post}} \cdot e^{-i \frac{\theta_i}{2} P} \cdot U_{\mathrm{pre}}|\psi\rangle
\end{equation}
\par
The gradient of $\Theta$ can be derived as follows:
\begin{equation}
    \begin{array}{cc}
         &  \frac{\partial f}{\partial\theta_i} = \frac{1}{2} \left[f(\Theta+\frac{\pi}{2}e_i)  - f(\Theta - \frac{\pi}{2}e_i) \right] \\
    \end{array}
\end{equation}
\par
where $e_i$ is a unit vector.

\begin{table*}[!tb]
\centering
\caption{The statistics of datasets used in the experiments (window size is 100)}
\label{tab:dataset}
\begin{tblr}{
  cell{1}{1} = {r=2}{},
  cell{1}{2} = {r=2}{},
  cell{1}{3} = {r=2}{},
  cell{1}{4} = {r=2}{},
  cell{1}{5} = {c=2}{},
  cell{1}{7} = {c=2}{},
  hline{1,7} = {-}{0.08em},
  hline{2} = {5-9}{},
  hline{3} = {-}{},
  hline{6} = {-}{},
}
Dataset      & Total Logs & Log Events & Total Seqs~~ & Training set (80\%)    &           & Testing set (20\%)           &            \\
             &            &            &              & \# Normal              & \# Anomaly          & \# Normal          & \# Anomaly \\
BGL          & 4,747,963  & 1,847      & 47,135        & 37,708                 & 4,009 (10.6\%)      & 9,427              & 817 (8.7\%)        \\
Spirit       & 5,000,000  & 2,880      & 50,000       & 40,000                 & 19,384 (48.5\%)     & 10,000             & 346 (3.5\%)        \\
Thunderbird  & 9,959,160  & 4,992      & 99,593       & 79,674                 & 816 (1.0\%)         & 19,919             & 27 (0.1\%)      
\end{tblr}
\end{table*}

\section{Dataset \& Metric}

\subsection{Dataset} \label{sec:dataset}
To evaluate the training performance of quantum machine learning compared to traditional machine learning for log anomaly detection, we selected three datasets: BGL, Spirit, and Thunderbird \cite{zhu2023loghub}. Table \ref{tab:dataset} summarizes the statistics of datasets used in our experiments.

\begin{itemize}
    \item \textbf{BGL} is an open dataset of logs collected from a BlueGene/L supercomputer system at Lawrence Livermore National Laboratory (LLNL) in Livermore, California, featuring 131,072 processors and 32,768 GB of memory. The log includes both alert and non-alert messages, which are distinguished by alert category tags. In the first column of the log, a "-" indicates non-alert messages, while all other entries represent alert messages. The label information is suitable for research in alert detection and prediction. This dataset has been employed in various studies on log parsing, anomaly detection, and failure prediction.
    \item \textbf{Spirit} is an open dataset of logs collected from the supercomputer system known as Spirit (ICC2) at Sandia National Laboratories (SNL) in Albuquerque, New Mexico. The system features 1,028 processors and 1,024 GB of memory. The log dataset encompasses various alert and non-alert messages, which have been appropriately categorized and tagged. In the log files, alert messages are designated by specific category tags, whereas non-alert messages are marked with a ``-" in the first column. This labeling scheme makes the dataset well-suited for research in log parsing, anomaly detection, and failure prediction. The Spirit dataset has been widely used in multiple studies to analyze and understand the behavior of supercomputer systems, offering valuable insights into system reliability, performance, and the nature of failures within such large-scale systems.
    \item \textbf{Thunderbird} is an open dataset of logs collected from a Thunderbird supercomputer system at Sandia National Laboratories (SNL) in Albuquerque, featuring 9,024 processors and 27,072 GB of memory. The log includes both alert and non-alert messages, which are identified by alert category tags. In the first column of the log, a ``-" indicates non-alert messages, while all other entries represent alert messages. The label information is suitable for research in alert detection and prediction.
\end{itemize}

\subsection{Evaluation Metrics}

To evaluate the performance of traditional machine learning models and quantum machine learning models across different datasets, we used the metrics as follows:

\begin{itemize}
    \item \textbf{Precision} The proportion of true positive samples among those predicted as positive by the model. A high precision indicates that the model's positive predictions are accurate, with relatively few false positives (FP). $Precision=\frac{TP}{TP+FP}$
    \item \textbf{Recall}  The proportion of actual positive samples that the model correctly identifies as positive. A high recall indicates that there are relatively few false negatives (FN).  $Recall=\frac{TP}{TP+FN}$
    \item \textbf{Specificity} The proportion of actual negative samples that the model correctly identifies as negative. A high specificity suggests that there are relatively few false positives (FP). $Specificity=\frac{TN}{TN+FP}$
    \item \textbf{F1 score} The harmonic mean of Precision and Recall, is used to balance these two metrics. The F1 score is particularly valuable in scenarios where a trade-off between precision and recall is necessary. F1 score $=2\times \frac{Precision\times Recall}{Precision+Recall}$
\end{itemize}


\section{Research Questions}

In this section, we investigate the influence of various factors on the performance of three QML models, and designed the following research questions:
\par

\textbf{RQ1: How do the QML models perform compared to classical models?}  

This study aims to explore the potential of QML in log-based anomaly detection. To this end, we compare the performance of classical models described in Section \ref{sec:baseline} against their quantum counterparts using three public datasets described in Section \ref{sec:dataset}.
\par
In the experimental setting, a fixed-size sliding window (100 log messages) is applied to group the raw log data with chronological selection. Note that, we do not shuffle the generated log sequences in this strategy, which can guarantee that only historical logs are used in the training phase, and there are no future logs used in this phase \cite{}.  The quantum models are designed with the following: the 4-qubit PQCs are constructed with $R_x$ gates for data encoding. The single-qubit operations are implemented with RX gates, while entanglement between qubits is introduced via CNOT gates. The  Adam optimizer is used for training and the learning rate is $1\times e^{-4}$. 

\par

\textbf{RQ2: How do the QML models perform with different encoding methods?} 
To investigate the impact of encoding methods on the performance of QML models, we evaluated three common quantum encoding techniques: amplitude encoding and angle encoding (using $R_x$, $R_y$, and $R_z$ gates). Each encoding method translates classical log data into quantum states differently, which directly influences the expressivity and efficiency of the quantum model. The dataset preprocessing steps and experiment settings are the same as RQ1. 
\par

\textbf{RQ3: How do the QML models perform with different numbers of qubits?}
An increased number of qubits enables the encoding of more input features and allows the entangled quantum states to capture richer correlations between these features. However, current quantum hardware imposes practical limitations on the number of qubits, and increasing qubit count risks exceeding hardware capabilities, which may result in operational failures. Furthermore, a larger number of qubits necessitates more complex quantum state initialization procedures and more robust error correction mechanisms to ensure the reliability of computations. To investigate these trade-offs, we evaluated the performance of QML models with varying numbers of qubits: $\{4, 6, 8\}$. 
\par

\begin{figure*}[!tb]
    \centering
    \begin{minipage}{0.24\textwidth}
        \centering
        \includegraphics[width=\textwidth]{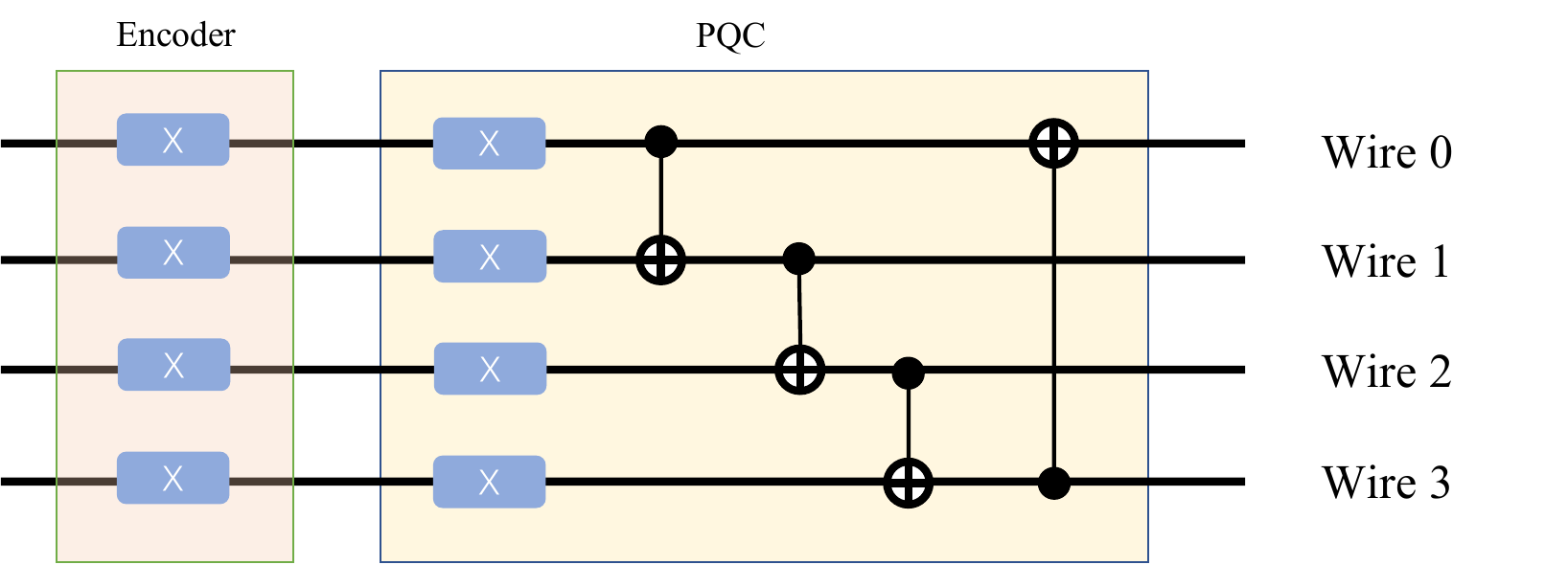}
        \text{$R_x$-based}  
    \end{minipage}%
    \hfill
    \begin{minipage}{0.24\textwidth}
        \centering
        \includegraphics[width=\textwidth]{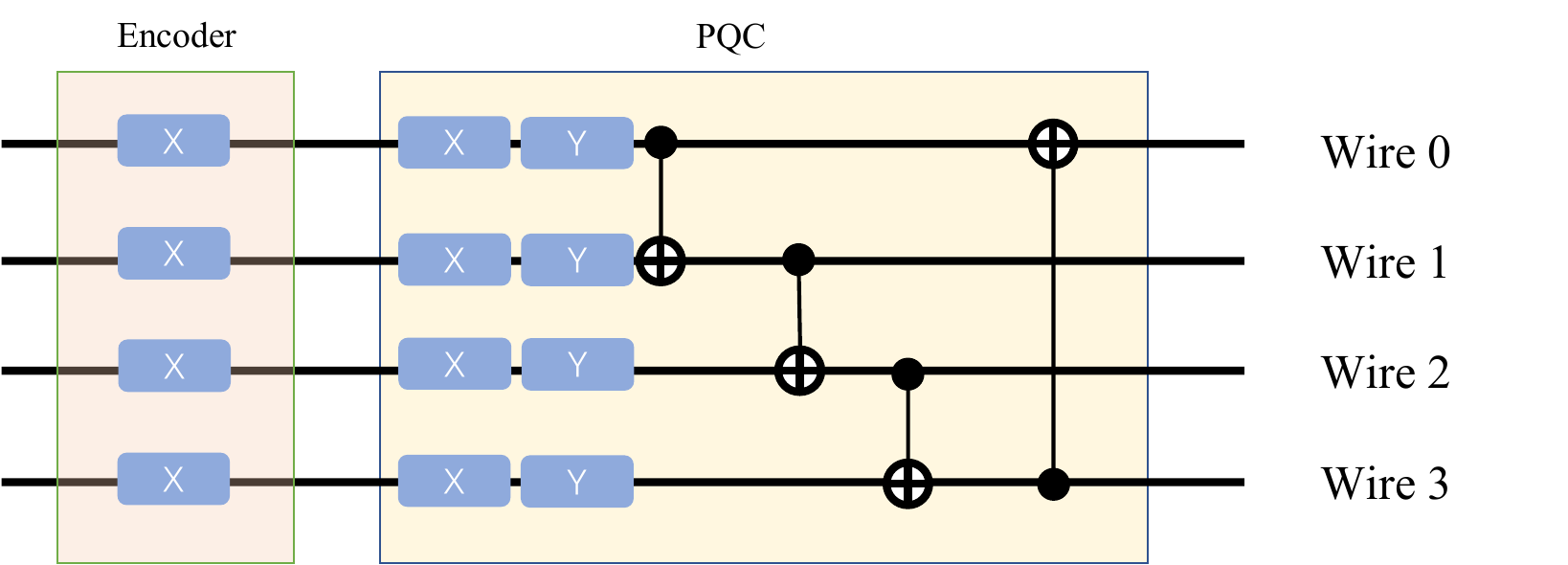}
        \text{$R_xR_y$-based}  
    \end{minipage}%
    \hfill
    \begin{minipage}{0.24\textwidth}
        \centering
        \includegraphics[width=\textwidth]{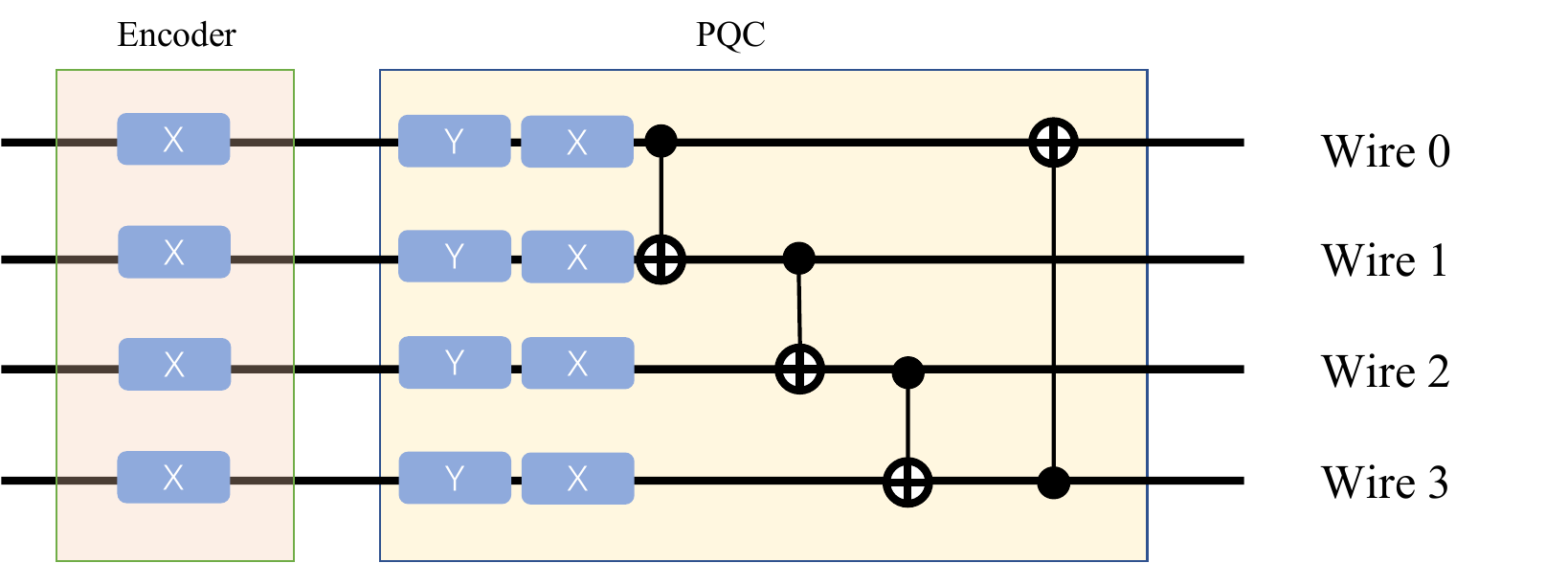}
        \text{$R_yR_x$-based}  
    \end{minipage}%
    \hfill
    \begin{minipage}{0.24\textwidth}
        \centering
        \includegraphics[width=\textwidth]{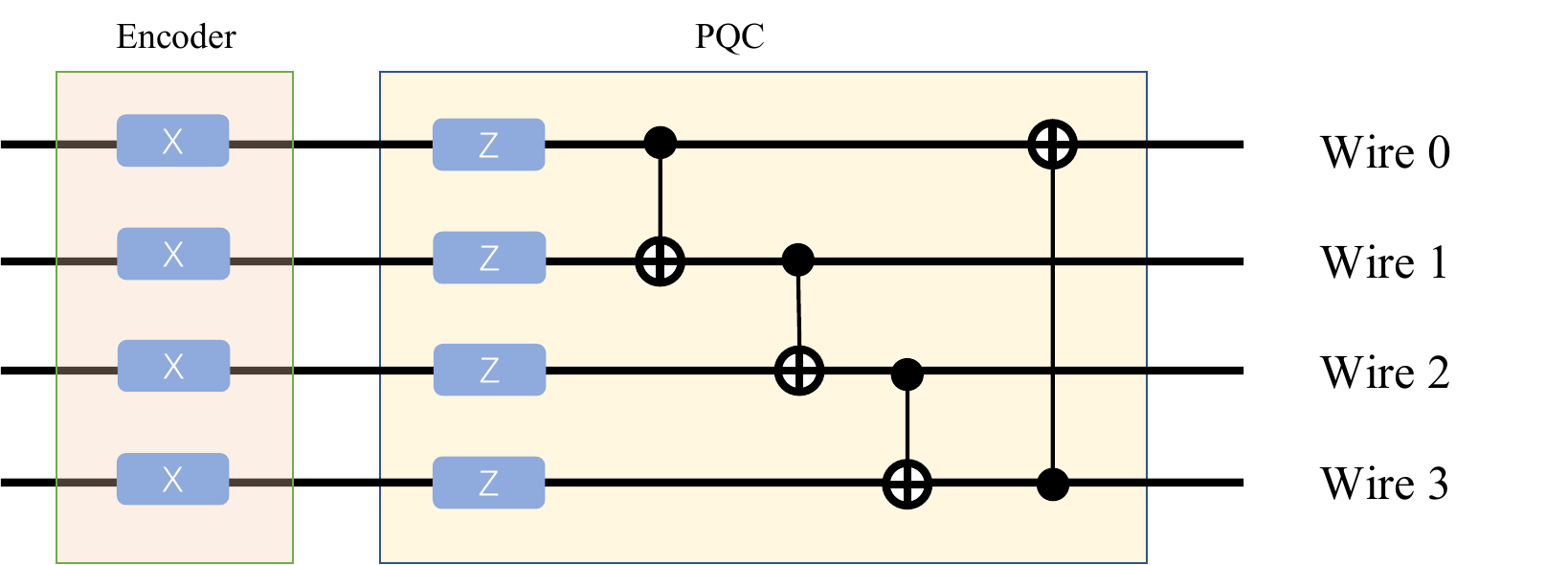}
        \text{$R_z$-based}  
    \end{minipage}
    \caption{The quantum circuit diagrams include: $R_x$, $R_xR_y$, $R_yR_x$, and $R_z$}
    \label{fig:pqc_design}
\end{figure*}

\textbf{RQ4: How do the QML models perform with different PQC designs?} 
This research question aims to evaluate the impact of different PQC designs on the performance of QML models for log-based anomaly detection. The structure and choice of quantum gates in a PQC influence its ability to model complex patterns and capture correlations in encoded data. To investigate this, we implemented and evaluated three PQC designs using different types of parameterized rotation gates: $R_x$, $R_xR_y$, $R_yR_x$, and $R_z$. Single-qubit operations were implemented using the aforementioned gates for parameterized rotations, combined with CNOT gates for entanglement. The $R_xR_y$ gate and $R_y R_x$ gate alternate between $R_x$ and $R_y$ gates for single-qubit operations while maintaining entanglement with CNOT gates. The four quantum circuit diagrams mentioned above are shown in Figure \ref{fig:pqc_design}.
\par
\textbf{RQ5: How do the QML models perform with different training set ratios?} 
We investigate the impact of varying training set sizes on the performance of models for log-based anomaly detection, particularly the QML model. Quantum computing enables the mapping of classical data to high-dimensional Hilbert space, where quantum states can leverage superposition and entanglement to enhance the separability of data, especially in small training sets. This efficient feature mapping allows the model to capture more expressive features in higher-dimensional spaces, which can improve performance in scenarios with limited data. In this experiment, we evaluated the QML models on subsets of the dataset with training set ratios of \(\{0.01, 0.1, 0.5, 1.0\}\), starting from the original 80\% training set. The same experiment settings are used described as in RQ1.
\par
\textbf{RQ6: How do the QML models perform in training efficiency?} 
Typically, a well-designed QML model will exhibit smooth and stable loss convergence, indicating that the learning process is stable and efficient. However, problems such as complex quantum circuits or suboptimal parameter initialization can lead to slow convergence or oscillations. Understanding whether the training process converges is important to verify the effectiveness and stability of the model, and can guide improvements in quantum circuit design and training strategies.

\begin{table*}[!tb]
\centering
\caption{Performance comparison on three baselines between quantum and classical counterparts}
\label{tab:main_results}
\begin{tabular}{ccccclccc} 
\toprule
\multirow{2}{*}{\textbf{Dataset }} & \multirow{2}{*}{\textbf{Metrics }} & \multicolumn{3}{c}{\textbf{Quantum }} &  & \multicolumn{3}{c}{\textbf{Classical}}  \\ 
\cmidrule{3-5}\cmidrule{7-9}
                                   &                                    & QDeepLog & QLogAnomaly & QLogRobust   &  & DeepLog & LogAnomaly & LogRobust         \\ 
\cmidrule{1-5}\cmidrule{7-9}
\multirow{4}{*}{BGL}               & Prec                               & 0.267~   & 0.254~      & 0.174~       &  & 0.271~  & 0.315~     & 0.938~            \\
                                   & Rec                                & 0.930~   & 0.988~      & 1.000~       &  & 0.988~  & 0.982~     & 0.977~            \\
                                   & Spec                               & 0.460~   & 0.386~      & 0.000~       &  & 0.438~  & 0.549~     & 0.986~            \\
                                   & F1                                 & 0.415~   & 0.404~      & 0.297~       &  & 0.425~  & 0.477~     & 0.957~            \\ 
\cmidrule{1-5}\cmidrule{7-9}
\multirow{4}{*}{Spirit}            & Prec                               & 0.415~   & 0.413~      & 0.413~       &  & 0.413~  & 0.507~     & 0.973~            \\
                                   & Rec                                & 1.000~   & 1.000~      & 1.000~       &  & 1.000~  & 1.000~     & 1.000~            \\
                                   & Spec                               & 0.010~   & 0.000~      & 0.000~       &  & 0.000~  & 0.317~     & 0.980~            \\
                                   & F1                                 & 0.587~   & 0.584~      & 0.584~       &  & 0.584~  & 0.673~     & 0.986~            \\ 
\cmidrule{1-5}\cmidrule{7-9}
\multirow{4}{*}{Thunderbird}       & Prec                               & 0.200~   & 0.200~      & 0.200~       &  & 0.211~  & 0.200~     & 0.889~            \\
                                   & Rec                                & 1.000~   & 1.000~      & 1.000~       &  & 1.000~  & 1.000~     & 1.000~            \\
                                   & Spec                               & 0.000~   & 0.000~      & 0.000~       &  & 0.062~  & 0.000~     & 0.969~            \\
                                   & F1                                 & 0.333~   & 0.333~      & 0.333~       &  & 0.348~  & 0.333~     & 0.941~            \\
\bottomrule
\end{tabular}
\end{table*}

\section{Results \& Analysis}
\subsection{Implementation Details.}

In this study, we evaluate the performance of DeepLog, LogAnomaly, LogRobust, and their quantumized counterparts on the BGL, Spirit, and Thunderbird datasets. Based on their open-source implementations and the description in Section III-B, the classical models were quantumized to develop QDeepLog, QLogAnomaly, and QLogRobust. Specifically, the PyTorch-based open-source implementations of DeepLog, LogAnomaly, and LogRobust are utilized, with hyperparameters tuned to achieve the best results.
\par
For QDeepLog and QLogAnomaly, the LSTM components in the original code are replaced with quantum-simulated QLSTM. The implementation is carried out using the torchquantum library \cite{hanruiwang2022quantumnas}, which by default employs 4 qubit quantum circuits. QLSTM first converts the data into quantum circuit inputs, which are processed by an encoder and subsequently fed into the PQC. The PQC comprises four QLayers, named input, update, forget, and output QLayer, each associated with its quantum gate circuit. These gates serve as trainable parameters for the subsequent model training. For QLogRobust, the attention mechanism is replaced with a quantumized implementation. Similarly, the default configuration uses an 8 qubit quantum circuit for QLayer. When computing $K$, $Q$, and $V$, the traditional linear layers are replaced with QLayer implementations.
\par
It is worth noting that the quantum circuits used in this study are simulated under ideal conditions, i.e. noise-free, and interference-free.

\subsection{RQ1: Performance on different log datasets}

To compare the differences between classical methods and their quantumized counterparts, this study evaluated the performance of three methods across three datasets. For each method type, the best-performing parameters were selected, and the experimental results are presented in Table \ref{tab:main_results}.
\par
We find that, on the Spirit dataset, the quantumized DeepLog (QDeepLog) method achieves a slightly higher F1 score than the classical DeepLog method. In contrast, the quantumized LogAnomaly (QLogAnomaly) model performs worse than its classical counterpart. On the BGL dataset, both the quantumized DeepLog and LogAnomaly models exhibit inferior performance compared to the classical methods. LogRobust, as a supervised model employing an attention mechanism, demonstrates significantly worse performance in its quantumized counterpart (QLogRobust) compared to its classical counterpart, primarily due to the excessive complexity of the quantum circuits. QLogRobust cascades two PQC circuits, LSTM and Attention. The quantum superposition and entanglement operations are increased, introducing more errors. Notably, on the Thunderbird dataset, all three quantumized methods achieve identical F1 scores, which may be attributed to the unique characteristics of this dataset. For example, Thunderbird contains the largest number of log events of the three datasets (4992), and the proportion of anomalies in the training set is only 1\%. 
\par

Our experimental results demonstrate the potential of QML models in log anomaly detection. However, the advantages of QML cannot be fully demonstrated by simply quantumizing existing classical models. For complex classical models, such as LogRobust, the performance is significantly reduced after quantumization. Therefore, among other RQs, we will mainly evaluate the QDeepLog and QLogAnomaly.

\begin{tcolorbox}[colback=gray!10,
                colframe=black,
                arc=2mm, 
                auto outer arc, 
                left=2mm, right=2mm,
                top=0.2mm, bottom=0.2mm
               ]
\textbf{Summary.} QML models do not consistently achieve superior performance over their classical counterparts. Performance differences observed in classical models do not generalize to quantum models. Performance varies depending on dataset characteristics, and greater complexity in quantum circuits may adversely affect performance.
\end{tcolorbox}

\par

\subsection{RQ2: The impact of different qubit numbers in quantum circuits}
In RQ1, we use the 4-qubits PQC to quantumized three classical methods. Because of the poor performance of QLogRobust and the difficulty of the Thunderbird dataset, in this RQ, the experiment was conducted with quantum circuits containing 4, 6, and 8 qubits on BGL to investigate the effect of the number of qubits counts on the performance of QDeepLog and QLogAnomaly. Table \ref{tab:n_qubits} shows the results.

\par
For both QDeepLog and QLogAnomaly, increasing the number of qubits does not consistently improve performance across metrics. For instance, in QDeepLog, precision slightly drops from 0.267 with 4 qubits to 0.251 with 6 qubits, before rising marginally to 0.258 with 8 qubits. This inconsistency suggests that the added qubits do not directly translate to better precision. Similarly, while recall remains relatively stable and high (e.g., 0.930 for 4 qubits and 0.924 for 8 qubits in QDeepLog), the low and variable specificity (e.g., 0.460 with 4 qubits versus 0.428 with 6 qubits) highlights the models’ struggles in reducing false positives. This may result from the increasing circuit complexity as the number of qubits increases, leading to potential overfitting or inefficiency in utilizing the additional qubits. Furthermore, the limited changes in F1 scores, such as QLogAnomaly’s slight rise from 0.352 with 4 qubits to 0.404 with 8 qubits, suggest diminishing returns in balancing precision and recall as circuits scale. 
\par
Our results illustrate that merely increasing the number of qubits does not ensure improved performance, emphasizing the need for careful quantum circuit design tailored to the task at hand. This trade-off between model complexity and performance highlights a potential overfitting issue or inefficiency in leveraging additional qubits. Furthermore, the limited improvement in F1 scores with higher qubit counts suggests diminishing returns in predictive performance as circuit complexity increases.

\begin{tcolorbox}[colback=gray!10,
                colframe=black,
                arc=2mm, 
                auto outer arc, 
                left=2mm, right=2mm,
                top=0.2mm, bottom=0.2mm
               ]
\textbf{Summary.} Increasing the number of qubits does not guarantee better performance. The obstacle to applying QML models to log-based anomaly detection is not the limitation of the number of qubits, but other potential factors that need to be further considered.
\end{tcolorbox}

\begin{table}[!tb]
\centering
\caption{Performance on BGL dataset with different numbers of qubits in the circuit}
\label{tab:n_qubits}
\begin{tabular}{ccccc} 
\toprule
\multirow{2}{*}{\textbf{Model}}       & \multirow{2}{*}{\textbf{Metrics}} & \multicolumn{3}{c}{\textbf{\# qubits}}  \\ 
\cmidrule{3-5}
                             &                          & 4      & 6      & 8            \\ 
\midrule
\multirow{4}{*}{QDeepLog}    & Prec                     & 0.267~ & 0.251~ & 0.258~       \\
                             & Rec                      & 0.930~ & 0.906~ & 0.924~       \\
                             & Spec                     & 0.460~ & 0.428~ & 0.440~       \\
                             & F1                       & 0.415~ & 0.393~ & 0.404~       \\ 
\midrule
\multirow{4}{*}{QLogAnomaly} & Prec                     & 0.214~ & 0.248~ & 0.254~       \\
                             & Rec                      & 0.994~ & 0.994~ & 0.988~       \\
                             & Spec                     & 0.230~ & 0.363~ & 0.386~       \\
                             & F1                       & 0.352~ & 0.397~ & 0.404~       \\
\bottomrule
\end{tabular}
\end{table}

\subsection{RQ3: The impact of different quantum encoding methods on model performance}

\begin{table}
\centering
\caption{Comparison of the performance of different encoding methods on the BGL dataset}
\begin{tblr}{
  cells = {c},
  cell{1}{1} = {r=2}{},
  cell{1}{2} = {r=2}{},
  cell{1}{3} = {c=3}{},
  cell{1}{6} = {r=2}{},
  hline{1,11} = {-}{0.08em},
  hline{2} = {3-5}{0.03em},
  hline{3,7} = {-}{0.05em},
}
\textbf{Model} & \textbf{Metrics} & \textbf{Angle} &       &       & \textbf{ Amplitude} \\
               &                  & RX                     & RY    & RZ    &                        \\
QDeepLog       & Prec             & 0.267                  & 0.220 & 0.245 & 0.211                  \\
               & Rec              & 0.930                  & 0.912 & 0.942 & 0.994                  \\
               & Spec             & 0.460                  & 0.319 & 0.389 & 0.216                  \\
               & F1               & 0.415                  & 0.355 & 0.389 & 0.348                  \\
QLogAnomaly    & Prec             & 0.254                  & 0.264 & 0.272 & 0.198                  \\
               & Rec              & 0.988                  & 0.988 & 0.994 & 0.988                  \\
               & Spec             & 0.386                  & 0.420 & 0.437 & 0.153                  \\
               & F1               & 0.404                  & 0.417 & 0.427 & 0.329                  
\end{tblr}
\end{table}

To explore the patterns associated with RQ3, the performance of QDeepLog and QLogAnomaly was evaluated on the BGL dataset using different encoding methods. Specifically, three types of angle encodings including $R_x$, $R_y$, and $R_z$ are initially applied during training, followed by an evaluation using amplitude encoding. The results are shown in Table IV.

For QDeepLog, $R_x$ angle encoding yielded the highest F1 score (0.415), followed by $R_z$ (0.389), while $R_y$ demonstrated the lowest performance (0.355) among the three angle encoding methods. Although amplitude encoding improved recall, it caused a significant decline in specificity, dropping from 0.460 with $R_x$ encoding to 0.216, a reduction of approximately 50\%. A similar pattern was observed with QLogAnomaly, where angle encoding consistently outperformed amplitude encoding. When $R_z$ encoding was used, the model achieved the highest F1 score (0.427), whereas amplitude encoding resulted in a lower F1 score of 0.329. The disparity was particularly pronounced in specificity: $R_z$ encoding achieved 0.437, while amplitude encoding attained only 0.153, approximately one-third of the specificity achieved by $R_z$ encoding.
 \par
The results highlight the significant influence of encoding methods on model performance. Angle encoding methods consistently produced better results, although the optimal angle varied between models. 

\begin{tcolorbox}[colback=gray!10,
                colframe=black,
                arc=2mm, 
                auto outer arc, 
                left=2mm, right=2mm,
                top=0.2mm, bottom=0.2mm
               ]

\textbf{Summary.} Quantum encoding significantly influence the performance. Angle encoding consistently outperforms amplitude encoding, with different angle encodings demonstrating optimal suitability for different models.
\end{tcolorbox}

\subsection{RQ4: The impact of different quantum circuit designs on model performance}

To answer RQ4, we evaluate QDeepLog and QLogAnomaly designed with different quantum circuits using four quantum gates, including $R_x$, $R_{xy}$, $R_{yx}$ and $R_z$. First, the design used only $R_x$ and CNOT gates for training. Next, an $R_y$ gate was added after each $R_x$ gate in the quantum circuit. Finally, the order of the $R_y$ and $R_x$ gates was reversed. The results are shown in Table \ref{tab:pqc}.

\par
For QDeepLog, the $R_x$-based circuit achieves the best overall performance, with the highest F1 score. Adding $R_y$ gates in the $R_xR_y$ design slightly increases recall (0.936 compared to 0.930 for $R_x$) but reduces precision and specificity, indicating a trade-off between detecting anomalies and minimizing false positives. Reversing the gate order in the $R_yR_x$ design further lowers performance, as seen in the reduced F1 score (0.388), reflecting sensitivity to gate sequence. For QLogAnomaly, the $R_z$-based circuit outperforms the others, achieving the highest performance across all metrics. This suggests that $R_z$ gates may enhance the model’s ability to discriminate between normal and anomalous logs. In contrast, the $R_xR_y$ circuit exhibits the poorest performance, with the lowest precision (0.206) and F1 score (0.398), possibly due to increased circuit complexity leading to overfitting or inefficient parameter utilization.
\par
These results emphasize that quantum circuit design has a significant impact on model performance. While adding gate complexity, as in $R_xR_y$ or $R_yR_x$ designs, can sometimes improve recall, it can also introduce trade-offs that compromise overall effectiveness. Therefore, the choice of the appropriate gate configuration is critical.

\begin{tcolorbox}[colback=gray!10,
                colframe=black,
                arc=2mm, 
                auto outer arc, 
                left=2mm, right=2mm,
                top=0.2mm, bottom=0.2mm
               ]
\textbf{Summary.} Quantum circuit design significantly impacts performance. Simpler designs such as $R_x$ achieve balanced metrics, while $R_z$ improves specificity. Complex designs such as $R_xR_y$ may improve recall, but have the overfitting problem.
\end{tcolorbox}

\begin{table}
\centering
\caption{Comparison of performance of different quantum circuit designs}
\label{tab:pqc}
\begin{tabular}{cccccc} 
\toprule
\multirow{2}{*}{\textbf{Model }} & \multirow{2}{*}{\textbf{Metrics }} & \multicolumn{4}{c}{\textbf{PQC Design}}  \\ 
\cmidrule{3-6}
                                 &                                    & $R_x$     & $R_xR_y$   & $R_yR_x$   & $R_z$      \\ 
\midrule
\multirow{4}{*}{QDeepLog}        & Prec                               & 0.267~ & 0.258~ & 0.247~ & 0.252~  \\
                                 & Rec                                & 0.930~ & 0.936~ & 0.906~ & 0.942~  \\
                                 & Spec                               & 0.460~ & 0.432~ & 0.416~ & 0.410~  \\
                                 & F1                                 & 0.415~ & 0.405~ & 0.388~ & 0.398~  \\ 
\midrule
\multirow{4}{*}{QLogAnomaly}     & Prec                               & 0.254~ & 0.206~ & 0.228~ & 0.270~  \\
                                 & Rec                                & 0.988~ & 0.942~ & 0.994~ & 0.988~  \\
                                 & Spec                               & 0.386~ & 0.410~ & 0.290~ & 0.435~  \\
                                 & F1                                 & 0.404~ & 0.398~ & 0.371~ & 0.424~  \\
\bottomrule
\end{tabular}
\end{table}

\subsection{RQ5: The impact of different training set sizes on the performance of various models and their quantumized counterparts}

We evaluate the models studied with different training set sizes. Our default setting is to use 80\% of the log data as the training set and 20\% as the test set. In this setting, 50\%, 20\%, 10\%, and 1\% of the original training set are randomly sampled as the training set, and the test set is kept the same. Specifically, as in RQ2, we selected DeepLog and LogAnomaly and their quantumized counterparts for evaluation on BGL. As both models are based on semi-supervision, i.e. only normal logs are used for training, we remove the abnormal log sequences in the training set and sample them as the original training set. The results are shown in Table \ref{tab:train_ratio}.
\par
Generally, both classical and quantumized methods demonstrate robustness to reduced training ratios. For example, classical LogAnomaly maintains relatively stable precision (e.g. 0.315 at 100\% training ratio versus 0.302 at 1\%) and F1 scores (e.g. 0.477 at 100\% versus 0.461 at 1\%), indicating its robustness in semi-supervised learning scenarios. In contrast, QLogAnomaly experiences greater variability in precision, dropping from 0.254 at 100\% to 0.237 at 50\%, but recovering slightly to 0.257 at 1\%, reflecting sensitivity to training set size, especially at intermediate ratios. Interestingly, recall remains high and stable for all models, such as 0.988 for QLogAnomaly and 0.994 for DeepLog at various training ratios, suggesting that these models effectively capture normal patterns even with limited training data. However, specificity varies significantly, particularly for QDeepLog, which drops from 0.460 at 100\% training ratio to 0.340 at 10\%, highlighting the challenges of reducing false positives when the training set is smaller. Notably, the F1 scores for QDeepLog decline more steeply at reduced training ratios (e.g. 0.415 at 100\% to 0.377 at 10\%), suggesting that its performance balance deteriorates more compared to its classical counterpart. 
\par
These observations suggest that quantumized models are more sensitive to training data reductions compared to classical counterparts, potentially due to their reliance on more complex parameter spaces, which may require larger datasets to achieve optimal performance. Additionally, the stability of recall alongside declining specificity in quantumized models implies potential overfitting to normal patterns. 
\begin{tcolorbox}[colback=gray!10,
                colframe=black,
                arc=2mm, 
                auto outer arc, 
                left=2mm, right=2mm,
                top=0.2mm, bottom=0.2mm
               ]
\textbf{Summary.} Quantumized models are robust to training set size, with stable recall but fluctuating precision and specificity, reflecting challenges in generalization. 
\end{tcolorbox}

\begin{figure*}[!tb]
    \centering
    \includegraphics[scale=0.3]{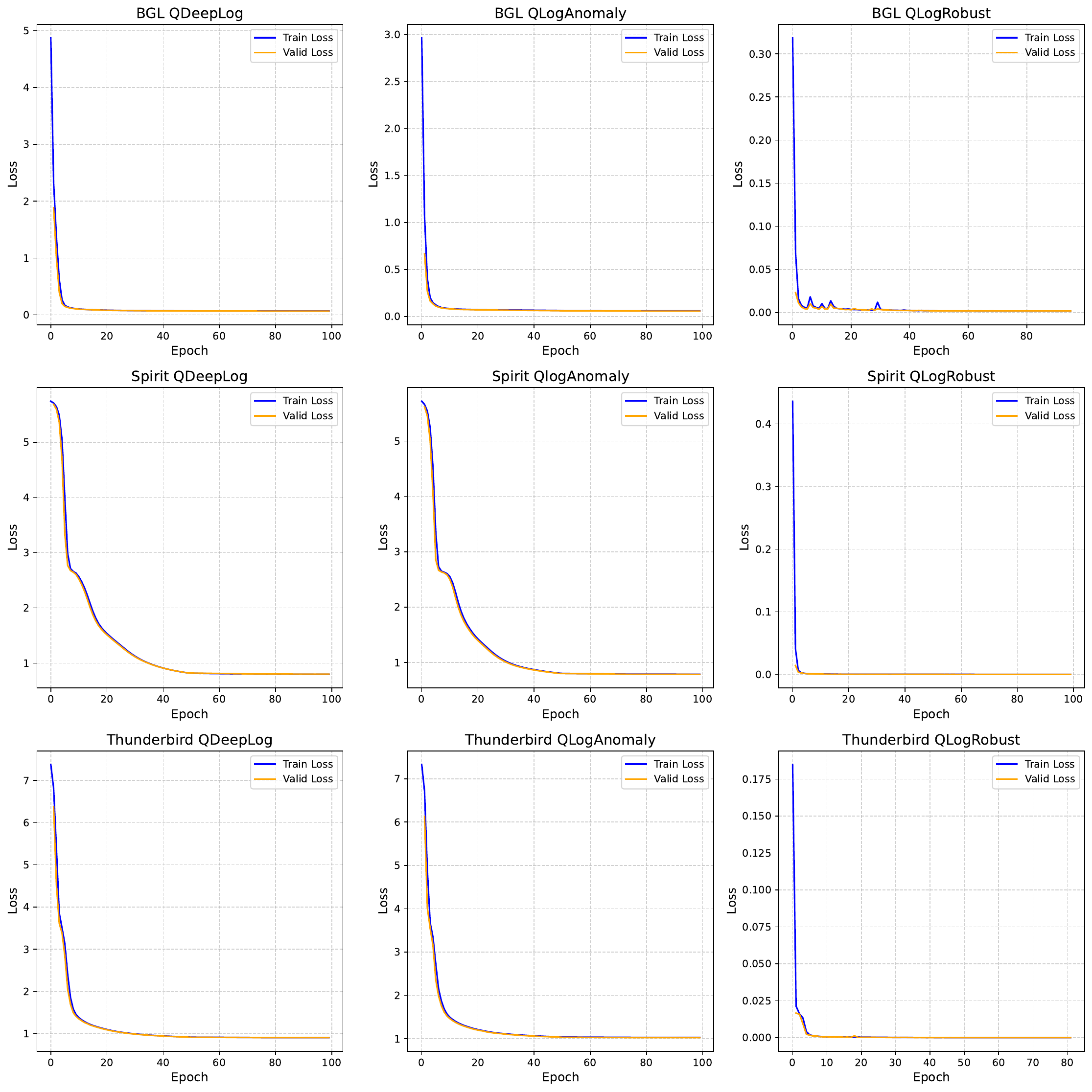}
    \caption{Training loss and validation loss curves on three datasets}
    \label{fig:loss}
\end{figure*}

\begin{table}
\centering
\caption{Performance comparison of different training ratios on BGL Dataset}
\label{tab:train_ratio}
\resizebox{1.0\linewidth}{!}{
\begin{tabular}{lllllll} 
\toprule
\multirow{2}{*}{\textbf{Model}}       & \multirow{2}{*}{\textbf{Metrics}} & \multicolumn{5}{c}{\textbf{Training Ratio}}          \\ 
\cmidrule{3-7}
                             &                          & 100\%      & 50\%    & 20\%    & 10\%    & 1\%    \\ 
\midrule
\multirow{4}{*}{LogAnomaly}  & Prec                     & 0.315~ & 0.313~ & 0.313~ & 0.315~ & 0.302~  \\
                             & Rec                      & 0.982~ & 0.977~ & 0.988~ & 0.982~ & 0.977~  \\
                             & Spec                     & 0.549~ & 0.547~ & 0.542~ & 0.549~ & 0.523~  \\
                             & F1                       & 0.477~ & 0.474~ & 0.475~ & 0.477~ & 0.461~  \\ 
\midrule
\multirow{4}{*}{QLogAnomaly} & Prec                     & 0.254~ & 0.237~ & 0.251~ & 0.264~ & 0.257~  \\
                             & Rec                      & 0.988~ & 0.988~ & 0.994~ & 0.994~ & 0.988~  \\
                             & Spec                     & 0.386~ & 0.327~ & 0.375~ & 0.414~ & 0.398~  \\
                             & F1                       & 0.404~ & 0.382~ & 0.401~ & 0.417~ & 0.408~  \\ 
\midrule
\multirow{4}{*}{DeepLog}     & Prec                     & 0.271~ & 0.270~ & 0.268~ & 0.245~ & 0.271~  \\
                             & Rec                      & 0.988~ & 0.988~ & 0.994~ & 0.994~ & 0.988~  \\
                             & Spec                     & 0.438~ & 0.437~ & 0.426~ & 0.353~ & 0.438~  \\
                             & F1                       & 0.425~ & 0.425~ & 0.422~ & 0.393~ & 0.425~  \\ 
\midrule
\multirow{4}{*}{QDeepLog}    & Prec                     & 0.267~ & 0.246~ & 0.264~ & 0.235~ & 0.268~  \\
                             & Rec                      & 0.930~ & 0.953~ & 0.930~ & 0.959~ & 0.924~  \\
                             & Spec                     & 0.460~ & 0.383~ & 0.453~ & 0.340~ & 0.468~  \\
                             & F1                       & 0.415~ & 0.391~ & 0.411~ & 0.377~ & 0.416~  \\
\bottomrule
\end{tabular}}
\end{table}
\begin{table}[!tb]
\centering
\caption{Comparison of the number of parameters of classical (with float32) and quantum models}
\label{tab:params}
\begin{tabular}{ccc} 
\toprule
\multirow{2}{*}{\textbf{Model }} & \multicolumn{2}{c}{\textbf{\# Params }}  \\ 
\cmidrule{2-3}
                                 & classical       & quantum                  \\ 
\midrule
\multicolumn{1}{l}{LSTM}         & 2,281,472 bit  & 8,896 bit + 16 qubit         \\
\multicolumn{1}{l}{Attention}    & 4,202,496 bit & 1,728 bit + 8 qubit          \\
\bottomrule
\end{tabular}
\end{table}

\subsection{RQ6: Training efficiency analysis}

To answer RQ6, we plot the training and validation loss curves of the model evaluated in RQ1 on three log datasets to evaluate the training efficiency of QML models. We set the maximum epoch to 100 and keep the other parameters the same as in RQ1. The results are shown in Figure \ref{fig:loss}.
\par

All three models show smooth and stable loss curves, and most of them reach convergence in training loss and validation loss after a small number of epochs (E$\leq20$), demonstrating the high efficiency of QML training in log anomaly detection. We observe that QLogRobust has slight oscillations on the BGL and Thunderbird datasets, which may be due to the complexity of the QLogRobust quantum circuit and the characteristics of the datasets (the proportion of anomalies in the training set is small, ~10\% and ~1\%, respectively). In addition, the QDeepLog and QLogAnoamaly models converge slowly on Spirit, which requires further exploration of the data and model aspects.
\par

Finally, we calculate the parameter sizes for LSTM and Attention models (see Table \ref{tab:params}). For LSTM, the classical model requires 2,281,472 bits of storage, while the QLSTM model only requires 8,896 bits and 16 quantum qubits. For Attention, the classical model occupies 4,202,496 bits, whereas the QAttention model requires just 1,728 bits and 8 quantum qubits. These results demonstrate that quantumized models significantly reduce parameter size compared to their classical counterparts, and can have lower energy consumption.

\begin{tcolorbox}[colback=gray!10,
                colframe=black,
                arc=2mm, 
                auto outer arc, 
                left=2mm, right=2mm,
                top=0.2mm, bottom=0.2mm
               ]
\textbf{Summary.}  The QML model can effectively capture the potential patterns in the logs and ensure the stability of the training process. Meanwhile, the QML model has fewer parameters and theoretically less energy consumption.
\end{tcolorbox}

\section{Discussion}
Our study highlights the comparative strengths, limitations, and potential of QML versus classical ML for log-based anomaly detection. By evaluating QML models against classical counterparts such as DeepLog, LogAnomaly, and LogRobust, we have identified key findings that underscore the transformative potential of QML, while also highlighting areas that require further refinement. Based on our findings, we can conclude that quantumized existing classical models do not always perform better than their classical counterparts. The performance is affected by serval factors including log data characters, quantum encode method, quantum circuit design, and the number of qubits. 
\par
\subsection{Advantages}
\textbf{Parameter Efficiency and Complexity Reduction:}  QML models using PQCs can achieve comparable or better performance with fewer trainable parameters. This is particularly beneficial in resource-constrained environments where memory and computing power are limited. The inherent parallelism of quantum systems allows high-dimensional data to be represented more compactly than in classical models.
\par
\textbf{Enhanced Data Representations:} By operating in a high-dimensional Hilbert space, QML models can capture complex, non-linear patterns in logs. Quantum superposition and entanglement allow QML to explore larger solution spaces, potentially leading to better detection performance.
\par
\textbf{Potential for Improved Generalization:} QML models demonstrated robust recall across various training set sizes, indicating their capacity to learn normal patterns effectively even with limited data. This suggests that QML could be particularly useful in semi-supervised settings.

\subsection{Limitations}
\textbf{Performance Variability:} The current generation of QML models does not consistently outperform their classical counterparts. For complex datasets such as Thunderbird, classical models such as LogRobust retained superior accuracy and specificity. This variability suggests that QML, while promising, is highly sensitive to dataset characteristics and circuit design.
\par
\textbf{Quantum Circuit Complexity:}  More complex PQCs, such as those incorporating multiple rotation gates or cascading circuits, often lead to performance degradation. This suggests a risk of overfitting and computational inefficiencies. Simplifying quantum circuits and optimizing gate configurations remain critical challenges.
\par
\textbf{Hardware Constraints:} The simulations in our study assumed a perfect quantum computer $\textendash$ no noise (loss, decoherence, etc.), precise control, and perfect error correction. However, the scalability of QML models is limited by the limitations of the Noisy Intermediate-Scale Quantum (NISQ) device, particularly in terms of the number of qubits and error rates, i.e. maintaining performance without introducing errors becomes increasingly difficult when running on a real quantum computer. 

\subsection{Future work}
Future research will focus on improving the robustness and scalability of QML for LogAD. For example, developing optimized PQCs tailored to specific log data characteristics and exploring advanced quantum encoding strategies to improve data representation. In addition, it is a very interesting direction to explore noisy QML models, which would allow performance evaluation when running on NISQ devices.

\section{Conclusion}
In this paper, we first introduce quantum machine learning to log-based anomaly detection and design a unified framework to evaluate the performance of quantum machine learning models. We quantumize three commonly used classical models and perform extensive evaluations. Our results show that the performance of quantum machine learning models has potential, but this requires careful circuit design. We also suggest possible future works and hope our results can inspire researchers to explore quantum machine learning, a future computing paradigm, for log-based anomaly detection.


\bibliography{ref}
\bibliographystyle{IEEEtran}

\end{document}